\pdfoutput=1

\documentclass[11pt]{article}

\usepackage[]{EACL2023}
\usepackage{times}
\usepackage{latexsym}

\usepackage[T1]{fontenc}

\usepackage[utf8]{inputenc}

\usepackage{microtype}
\usepackage{amsmath}
\usepackage{graphicx}
\usepackage{multirow}
\usepackage{soul}

\usepackage{pifont}

\usepackage{times}
\usepackage{latexsym}
\usepackage{graphicx}
\usepackage{booktabs}
\usepackage{subcaption}
\usepackage[ruled,vlined]{algorithm2e}
\usepackage{multirow}
\usepackage{tabularx}
\usepackage{amsmath}
\usepackage{bbold}
\usepackage{mathtools}
\usepackage{xcolor}
\usepackage{booktabs}
\usepackage{tabularx,ragged2e}
\usepackage{enumitem}
\newcommand{\name}{\textsc{F}easibility\textsc{QA}\xspace}

\title{\textit{John is 50 years old, can his son be 65?} \\Evaluating NLP Models' Understanding of Feasibility}

\author
{Himanshu Gupta  \hspace{9pt}  Neeraj Varshney \hspace{9pt}  Swaroop Mishra  \hspace{9pt} Kuntal Kumar Pal \hspace{9pt}\\
\textbf{Saurabh Arjun Sawant} \hspace{9pt} \textbf{Kevin Scaria} \hspace{9pt} \textbf{Siddharth Goyal} \hspace{9pt} \textbf{Chitta Baral} \\
{Arizona State University}\\
}

\begin{document}
\maketitle
\begin{abstract}
In current NLP research, large-scale language models and their abilities are widely being discussed. Some recent works have also found notable failures of these models. Often these failure examples involve complex reasoning abilities. 
This work focuses on a simple commonsense ability, reasoning about when an action (or its effect) is feasible. 
To this end, we introduce \name{}, a question-answering dataset involving binary classification (BCQ) and multi-choice multi-correct questions (MCQ) that test understanding of feasibility. 
We show that even state-of-the-art models such as GPT-3, GPT-2, and T5 struggle to answer the feasibility questions correctly. Specifically, on MCQ and BCQ questions, GPT-3 achieves an accuracy of just (19\%, 62\%) and (25\%, 64\%) in zero-shot and few-shot settings, respectively. 
We also evaluate models by providing relevant knowledge statements required to answer the question. We find that the additional knowledge leads to a 7\% gain in performance, but the overall performance still remains low. 
These results make one wonder how much commonsense knowledge about action feasibility is encoded in state-of-the-art models and how well they can reason about it.
\footnote{Dataset, baseline approaches, and instruction-tuned modeling approaches are freely available at \url{https://github.com/kevinscaria/feasibilityQA}} 


\end{abstract}

\section{Introduction}

Commonsense reasoning has been a key aspect of AI since its advent in the 1950s. It is closely associated with reasoning about actions and their effects, which is considered a big challenge, especially for deep learning-based AI approaches and language models \cite{lecun2022path,Dalvi2018TrackingSC,banerjee2020can}. While several datasets have been developed to evaluate large-scale language models, in this paper, we propose a dataset focused on reasoning about actions and their effects; specifically, the ability to reason if an action or its effect is feasible.


\begin{figure}[t!]
    \centering
    \includegraphics[width=7cm]{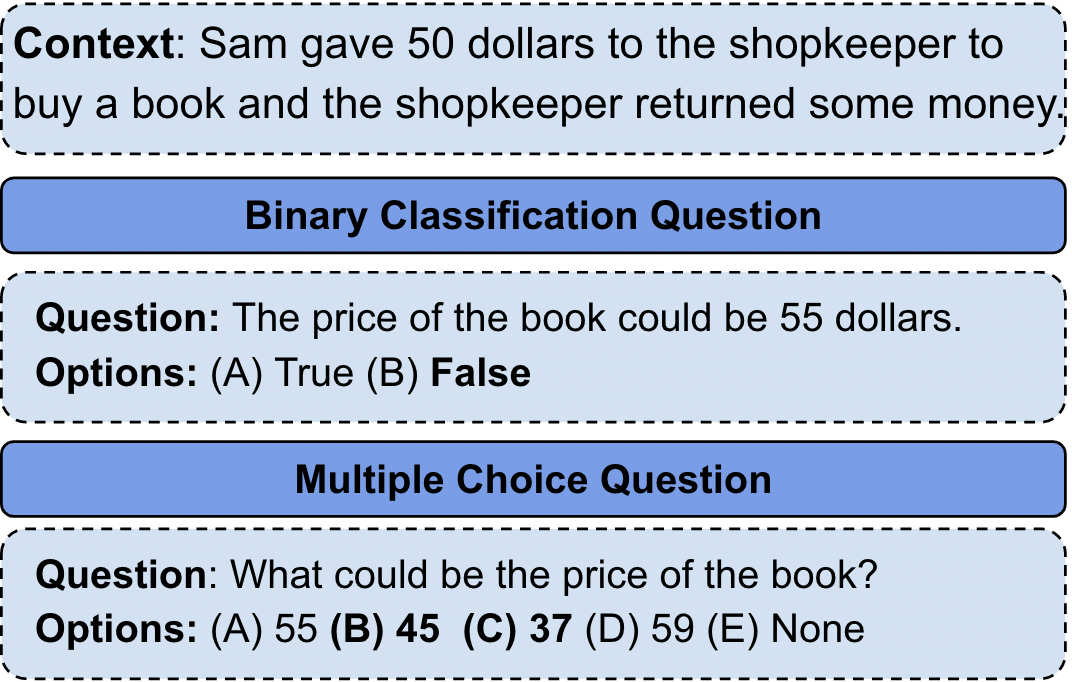}
    \caption{Illustrating a binary classification (BCQ) and a multiple choice question (MCQ) from \name{}. The correct answer options (\textbf{False} in BCQ and \textbf{(45, 37)} in MCQ) are highlighted in \textbf{bold}.} 
    \label{fig:teaser_figure}
\end{figure}


\begin{table*}[t!]
    \centering
    \small
    \resizebox{\linewidth}{!}
    {
    \begin{tabular}{llll}
\hline
\multicolumn{1}{c}{\textbf{Category}}                                                    & \multicolumn{1}{c}{\textbf{Knowledge}}                                                                        & \multicolumn{1}{c}{\textbf{Context}}                                                                                                                                           & \multicolumn{1}{c}{\textbf{Questions}}                                                                \\ \hline
\multirow{3}{*}{Attribute Comparison}                                                    & \multirow{3}{*}{\begin{tabular}[c]{@{}l@{}}Larger volume holds more \\ amount of water.\end{tabular}}         & \multirow{3}{*}{\begin{tabular}[c]{@{}l@{}}Barrett's has two cylindrical shaped \\ bottles. A bottle with a higher volume \\ holds 32 units of water.\end{tabular}}              & Amount of water in other bottle could be 28 units.                                                     \\
                                                                                         &                                                                                                               &                                                                                                                                                                                & Amount of water in other bottle could be 33 units.                                                     \\
                                                                                         &                                                                                                               &                                                                                                                                                                                & What could be the amount of water in other bottle?                                                    \\ \hline
\multirow{3}{*}{Change with Time}                                                        & \multirow{3}{*}{Age increases with time}                                                                      & \multirow{3}{*}{\begin{tabular}[c]{@{}l@{}}Today, while filling the \\ application form Edward \\ filled the age field with 16.\end{tabular}}                                  & Edward could have been 8 years old on his last birthday.                                               \\
                                                                                         &                                                                                                               &                                                                                                                                                                                & Edward could have been 19 years old on his last birthday.                                              \\
                                                                                         &                                                                                                               &                                                                                                                                                                                & What could have been Edward's age last year?                                                          \\ \hline
\multirow{3}{*}{Change with Action}                                                      & \multirow{3}{*}{\begin{tabular}[c]{@{}l@{}}Selling something reduces \\ its quantity\end{tabular}}            & \multirow{3}{*}{\begin{tabular}[c]{@{}l@{}}Joshua organized a garage sale \\ yesterday. Joshua sold a total of \\ 273 items at a minimum price of \\ 1 USD each.\end{tabular}} & Joshua could have made 300 dollars in the garage sale.                                                   \\
                                                                                         &                                                                                                               &                                                                                                                                                                                & Joshua could have made 260 dollars in the garage sale.                                                   \\
                                                                                         &                                                                                                               &                                                                                                                                                                                & \begin{tabular}[c]{@{}l@{}}How much money Joshua could have made \\ from the garage sale?\end{tabular} \\ \hline
\multirow{3}{*}{\begin{tabular}[c]{@{}l@{}}Implicit \\ Numerical Knowledge\end{tabular}} & \multirow{3}{*}{4 quarters make 1 dollar}                                                                     & \multirow{3}{*}{\begin{tabular}[c]{@{}l@{}}Christopher is accepting \\ quarters for a donation and \\ fails to collect 12 dollars.\end{tabular}}                               & He could have collected 35 quarters.                                                                   \\
                                                                                         &                                                                                                               &                                                                                                                                                                                & He could have collected 52 quarters.                                                                   \\
                                                                                         &                                                                                                               &                                                                                                                                                                                & How many quarters could Christopher have collected?                                                    \\ \hline
\multirow{3}{*}{Non Numerical}                                                           & \multirow{3}{*}{\begin{tabular}[c]{@{}l@{}}New movies can be watched \\ after the release date.\end{tabular}} & \multirow{3}{*}{\begin{tabular}[c]{@{}l@{}}The latest superhero movie was \\ releasing on 28th February 2022.  \\ Ashton wanted to see the movie.\end{tabular}}               & He could have watched the movie on 2nd March 2022.                                                     \\
                                                                                         &                                                                                                               &                                                                                                                                                                                & He could have watched the movie on 3rd February 2022.                                                  \\
                                                                                         &                                                                                                               &                                                                                                                                                                                & When could Ashton have watched the movie ?                                                             \\ \hline
\end{tabular}
    }
    \caption{
    Illustrative examples of two binary choice questions and one multiple choice question corresponding to a context statement. We also provide the corresponding category and relevant knowledge for the questions.
    }
    \label{tab:dataset_example}
\end{table*}

Figure \ref{fig:teaser_figure} illustrates an example of our dataset; given the information ``\textit{Sam gave 50 dollars to the shopkeeper to buy a book and the shopkeeper returned some money}'', it is not possible to compute the exact price of the book; however, it can be established that the feasible price of the book is \textit{less than} 50 since the shopkeeper returned some money. 
We often come across such questions in our daily life and find it trivial to reason about them.
Therefore, in order to develop NLP systems that can reliably reason about real-world situations, it is important to evaluate their understanding of feasibility.

Recently, many datasets have been created that test different reasoning skills such as pronoun resolution \cite{sakaguchi2021winogrande,levesque_winograd_2012}, commonsense reasoning \cite{singh-etal-2021-com2sense,Mihaylov2018CanAS,banerjee2021commonsense}, numerical reasoning \cite{Mishra2022LilaAU,ravichander-etal-2019-equate, lin-etal-2020-birds, zhang-etal-2020-language-embeddings,amini-etal-2019-mathqa,mishra-etal-2022-numglue,creswell2022selection, Pal2021InvestigatingNL}, qualitative reasoning \cite{tafjord-etal-2019-quartz, tafjord2019quarel}, discrete reasoning \cite{dua-etal-2019-drop}, and temporal reasoning \cite{zhou-etal-2019-going}. 
However, they do not have an ample number of examples that test understanding of feasibility. 

In this work, we address the above limitation and introduce \name{}, a dataset consisting of questions that require an understanding of feasibility. This dataset comprises of two types of questions: binary classification (BCQ) and multi-choice multi-correct questions (MCQ). In BCQ, the task is to determine whether the question is feasible or not given a context; in MCQ, the task is to select all feasible answers to the given question. The dataset consists of $\sim$5K instances covering diverse aspects of feasibility. 
Table \ref{tab:dataset_example} illustrates examples of various categories of feasibility questions.


We conduct comprehensive experiments with GPT-3, GPT-2, and T5 models \cite{brown2020language,radford2019language,raffel2020exploring} in zero-shot and few-shot settings and show that all of these models struggle to correctly answer feasibility questions.
Specifically, on (MCQ, BCQ) questions, GPT-3 achieves an accuracy of just ($19\%$, $62\%$) and ($25\%$, $64\%$) in zero-shot and few-shot settings, respectively. 

Prior work has found that explicitly providing relevant knowledge helps the model reason better and improves its performance  \cite{chen2018natural,xiong-etal-2019-improving,banerjee-etal-2019-careful,varshney2022can}.
We explore this aspect of reasoning by explicitly providing relevant knowledge statements and find that it leads to $\sim$7\% improvement in performance. However, the overall performance still remains low.
We further investigate GPT-3's ability to reason about feasibility questions by prompting it to generate the reasoning chain. 
In many cases, we find that GPT-3 successfully generates the correct reasoning chain but still fails to output the correct answer.
This analysis further leads to several interesting findings (Section \ref{sec_experiments}).

\section{FeasibilityQA}

\subsection{Dataset Creation}
For creating data instances of \name{}, we first create a context statement that describes a real-life situation. 
Then, we write two binary classification questions and one multiple choice question conditioned on the context that tests the understanding of feasibility. 

\paragraph{Dataset creation and verification process}
\label{dataset_verification_processs}
Seven computer science graduate students were involved in creating the dataset. 
Dataset creation consists of 3 phases. 
First, in the data creation stage, each student created 700 samples 
over the period of 3 months. 
In the next phase, each dataset creator’s questions were verified by a different student to ensure fairness during data validation. 
The 3rd stage of the validation was done when all the questions were compiled and cross-verified.
In each verification stage, the dataset creators rejected some samples where the inter-annotator agreement was low.
\footnote{All the dataset creators are authors of the paper.}



\subsubsection{Context Creation}
\label{subsec_context_creation}
We create context statements from real-life situations spanning diverse topics such as elementary physics, profit-loss scenarios, temporal comparisons, and quantity comparisons. 
We divide the contexts into the following five categories:

\textbf{Attribute comparison:} This category consists of questions that test feasibility aspects involving the comparison of attributes of two quantities. 
\textbf{Implicit numerical:} Questions in this category involve fundamental mathematical facts that test the ability to use those facts in real-world situations. 
\textbf{Change with action:} This category tests the ability to perceive a change in an item or state as an outcome of an action. 
\textbf{Change with time:} Here, questions test the understanding of feasibility related to temporal-based events. 
\textbf{Non Numerical:} This category includes questions where numbers are not explicitly involved in reasoning about feasibility. 
Table \ref{tab:dataset_example} provides examples of these categories. 
More details about them are in Appendix~\ref{appendixa_dataset}.

\paragraph{Motivation behind category selection}
The motivation behind developing a large language model such as GPT-3 is to mimic human intelligence and come closer to Artificial General Intelligence. 
We attempt to gauge the performance of models' intelligence by  developing simple commonsense reasoning questions. 
GPT-3 models are few-shot learners but find it hard to do proper numerical reasoning. Earlier datasets like this attempted to analyze numerical reasoning in this aspect. We are also trying to study it in the aspect of feasibility. Previous datasets, such as Numersense~\cite{lin-etal-2020-birds} and MC Taco~\cite{zhou-etal-2019-going}, do not have such a category, and we tried to bridge those gaps.

We think that these five categories are also a good representation of numerical feasibility. We found that questions from those categories had an adequate amount of complexity that the average human could easily figure out. So we expected that large language models should also be able to understand and answer accordingly. We created these categories to compare the models' numerical reasoning ability with and without knowledge. This gives us insights into whether knowledge helps in each aspect. We hope that these comparative studies across these five preliminary categories will inspire more future categories. 


\paragraph{Target of our dataset:} Our selection of categories in feasibility is inspired by the limitations in existing datasets since it is not possible to cover all the aspects of feasibility exhaustively.

\begin{table}[t!]
\begin{tabular}{lrr}
\hline
\textbf{Category}    & \multicolumn{1}{l}{\textbf{\begin{tabular}[c]{@{}l@{}}Binary \\ Instances\end{tabular}}} & \multicolumn{1}{l}{\textbf{\begin{tabular}[c]{@{}l@{}}MCQ \\ Instances\end{tabular}}} \\ \hline
Attribute comparison & 1696                                                                                     & 848                                                                                   \\
Non numerical        & 700                                                                                      & 350                                                                                   \\
Implicit numerical   & 444                                                                                      & 222                                                                                   \\
Change with action   & 196                                                                                      & 98                                                                                    \\
Change with time     & 36                                                                                       & 18                                                                                    \\ \hline
Total                & 3072                                                                                     & 1536                                                                                  \\ \hline
\end{tabular}

\caption{
   Categorization of \name{} across different categories .
    }
    
\label{tab:dataset_distrib}
\end{table}
\begin{table}[t!]
\centering
\begin{tabular}{lr}
\hline
\centering
Correct Answers & \multicolumn{1}{l}{\#Instances} \\ \hline
1          & 555                                      \\
2        & 622                                      \\
3        & 285                                      \\
4    & 31                                       \\
None & 43                                       \\ \hline
\end{tabular}

\caption{
   Frequency of correct answers for MCQ section.
    }
    
\label{tab:number_of_answers}
\end{table}



    

\begin{table*}[t!]
    \centering
    \small
    \resizebox{\linewidth}{!}
    {
    \begin{tabular}{c|cccc|cccc|cccc}
\hline
          & \multicolumn{4}{c|}{\textbf{GPT-3}}                          & \multicolumn{4}{c|}{\textbf{GPT-2}}                        & \multicolumn{4}{c}{\textbf{T5}}                             \\ \cline{2-13} 
\textbf{} & \multicolumn{2}{c|}{BCQ (\%)}          & \multicolumn{2}{c|}{MCQ (\%)} & \multicolumn{2}{c|}{BCQ (\%)}        & \multicolumn{2}{c|}{MCQ (\%)} & \multicolumn{2}{c|}{BCQ (\%)}           & \multicolumn{2}{c}{MCQ (\%)} \\ \cline{2-13} 
          & w/o K & \multicolumn{1}{c|}{w/ K}    & w/o K       &w/ K          & w/o K & \multicolumn{1}{c|}{w/ K}  & w/o K       &  w/ K         & w/o K & \multicolumn{1}{c|}{w/ K}     & w/o K       & w/ K         \\ \hline
Zero Shot & 62.96  & \multicolumn{1}{c|}{69.11} & 19.43        & 25.89       & 50.00    & \multicolumn{1}{c|}{50.00} & 0.19        & 0.45       & 50.55 & \multicolumn{1}{c|}{50.64} & 0.13        & 0.39      \\
One Shot  & 57.94  & \multicolumn{1}{c|}{64.66} & 20.94        & 24.15       & 50.00    & \multicolumn{1}{c|}{50.00} & 0.58        & 1.69       & 50.61 & \multicolumn{1}{c|}{50.33} & 0.45        & 0.58      \\
Few Shot  & 64.72  & \multicolumn{1}{c|}{68.55} & 25.94        & 37.23       & 50.00    & \multicolumn{1}{c|}{50.00} & 0.97        & 0.39       & 49.81 & \multicolumn{1}{c|}{49.87} & 0.84        & 1.10      \\ \hline
\end{tabular}
    }
    \caption{
    \textit{Exact match accuracy} of three models in BCQ (Binary Classification) and MCQ (multi-choice multi-correct) tasks across three settings. w/o K  and  w/ K represents without knowledge and with knowledge respectively. 
    }
    \label{tab:bc_results}
\end{table*}
\subsubsection{Question Creation}
\label{subsec_question_creation}
From each context, we create two binary classification and one multiple-choice question. 
Recall that in our questions, the context may not provide sufficient information to find the exact answer. However, the information is sufficient to test the validity of question/answer options (notice the use of the word `\textbf{could}'). In question creation, we ensure that all our contexts and questions describe realistic situations, e.g., we keep a range of numerical entities and units appropriate for their context. Table \ref{tab:dataset_example} illustrates examples of our dataset.

\noindent\textbf{BCQ:} For each context, we create two binary classification questions where the correct answer is \textit{True} for one and \textit{False} for the other. Evaluating models' consistency in answering two contrasting hypotheses conditioned on the same context provides an additional benefit.

\noindent\textbf{MCQ:} For each context, we create a multi-correct multiple-choice question. Here, a context-question pair and the corresponding answer options are given, and the task is to select all feasible options for the question. We ensure that there is also a \textit{None} option, which needs to be selected when all the other options are not feasible.
For a question, \textit{one or more options} (including `None') could be correct.

\subsection{Dataset Statistics}

Our dataset consists of $1536$ contexts and $4608$ context-question pairs ($3072$ BCQ and $1536$ MCQ). 
The category-wise distribution of the dataset is present in Table \ref{tab:dataset_distrib}.
BCQ dataset is label balanced, i.e., $1536$ instances for each of \textit{True} and \textit{False} labels. MCQ dataset has a varying number of correct options. 
Table \ref{tab:number_of_answers} shows the number of correct answers in the MCQ category. 

\section{Experiments}
\label{sec_experiments}

\subsection{Performance Evaluation \& Metrics}
For \textbf{BCQ}, we calculate \textit{exact match accuracy} against the gold label (True or False) for each question.
We also use a stricter evaluation metric, \textit{pairwise accuracy}, to better estimate the model's capability of understanding feasibility. 
Here we only consider a sample to be correct if both BCQ (True and False questions) are correctly answered by the model for one context statement.
For \textbf{MCQ}, we evaluate \textit{exact match accuracy}, i.e., the model's prediction is considered to be correct if all the correct answer options are predicted. 
We also compute \textit{recall}, which we define as the number of correct answers predicted by the model out of all the correct answer options.
Other results (false positive, false negative, category-wise) are in Appendix \ref{appendixb_performance_results}. 



\paragraph{Models:} 
We evaluate the performance of \textbf{GPT-3} (Text-DaVinci-002, with 256 max tokens, top p of 1, and frequency \& presence penalty of 0), \textbf{T5-11B}, and \textbf{GPT-2 large} models on our dataset. 


     
    



\subsection{Results}

\paragraph{Low Performance of All Models:} Table \ref{tab:bc_results} shows the accuracy of all three models in zero-shot, one-shot, and few-shot settings. 
On BCQ, GPT-3 achieves \textit{exact match accuracy} of just $62.9\%$. 
GPT-2 and T5 perform even worse and achieve close to the majority baseline (50\%).
GPT-2 gets an exact 50, indicating that the model does not understand such feasibility reasoning \footnote{Please refer Appendix \ref{appendixb1_performance_results} for details}.
On MCQ, which is a more difficult task than BCQ, all models, including GPT-3, achieve a very low \textit{strict accuracy} score.
This highlights that feasibility questions are challenging for even state-of-the-art models. 

\begin{table}[t!]
\centering
\begin{tabular}{lrr}
\hline
\textbf{} & \multicolumn{1}{l}{w/o  Knowledge} & \multicolumn{1}{l}{w/  Knowledge} \\ \hline
Zero      & 46.2                                   & 49.9                                \\
One       & 63.5                                   & 64.5                                \\
Few       & 66.7                                   & 70.3                                 \\ \hline
\end{tabular}

\caption{
   \textit{Recall} scores of GPT-3 on MCQ task.
    }
    
\label{tab:recall}
\end{table}


\paragraph{Decrease in performance in one-shot setting:} 
In the one-shot setting, the model’s prediction is heavily influenced by the label of the example (one) presented to the model. 
This phenomenon is also observed in several prior zero-shot, and one-shot studies \cite{zhao2021calibrate}. 
A similar trend is observed in the chain of thought experiments (results described in Table \ref{tab:gpt3_w_exp}).

\noindent \textbf{Providing Knowledge Improves}  GPT-3's performance by $\sim7\%$ across all settings. The accuracy particularly increases ($\sim12\%$) in the MCQ task in the few-shot setting.
Although GPT-3 performs better than T5 and GPT-2, it achieves just $68.5\%$ and $37.2\%$ on BCQ and MCQs, respectively.

\paragraph{GPT-3 achieves High Recall Scores on MCQs:}
In Table \ref{tab:recall}, we show recall scores of GPT-3 on MCQs.
GPT-3 achieves a high score (up to 70\%), highlighting that it gives correct responses but fails to give all the correct responses.

\paragraph{Pairwise Evaluation:} 
Recently, instance-level analysis of the evaluation data has received considerable research attention \cite{zhong-etal-2021-larger,varshney-etal-2022-ildae,rodriguez-etal-2021-evaluation,Mishra2022HardnessOS}.
Motivated by this, we analyze GPT-3's performance on BCQ questions using the stricter \textit{pairwise accuracy} metric. Even though the model performs $\sim63\%$ using \textit{exact match accuracy}, Table \ref{tab:pairwise_results} shows that the models' performance is at most $\sim43\%$ via \textit{pairwise accuracy}, highlighting a performance gap. The accuracy increases ($\sim13\%$) when knowledge is introduced, and the gap between different settings also narrows down, indicating that the addition of knowledge helps. 

\begin{table}[t!]
\resizebox{\linewidth}{!}{
\begin{tabular}{lrr}
\hline
          & \multicolumn{1}{l}{w/o Knowledge} & \multicolumn{1}{l}{w/ Knowledge} \\ \hline
Zero Shot & 42.9                             & 56.8                              \\
One Shot  & 17.9                             & 34.3                              \\
Few Shot  & 39.8                             & 55.8                              \\ \hline
\end{tabular}
}

\caption{
    \textit{Pairwise Accuracy} of GPT-3 in BCQ Task.
    }
\label{tab:pairwise_results}
\end{table}





\paragraph{Category wise results:} 
Table \ref{tab:one_shot_category_results} shows the category-wise results in one-shot setting for BCQ and MCQ tasks on GPT-3. 
Non Numerical category performed the best out of all categories (15\% more than other categories). Addition of knowledge improves the performance of all categories by around 5\% to 10\%. Similar trends are observed across MCQ task as well. We believe that Non Numerical performed the best because these samples were conditioned around factual knowledge and that it could be present in the pretraining corpus of GPT-3. 
We further analyze this in Appendix \ref{appendixb_performance_results}.

\begin{table}[t!]
\resizebox{\linewidth}{!}{
\begin{tabular}{lrr|rr}
\hline
                     & \multicolumn{2}{c|}{\textbf{BCQ}}                     & \multicolumn{2}{c}{\textbf{MCQ}}                     \\ \cline{2-5} 
                     & \multicolumn{1}{l}{w/o K} & \multicolumn{1}{l|}{w/ K} & \multicolumn{1}{l}{w/o K} & \multicolumn{1}{l}{w/ K} \\ \hline
Attribute Comparison & 58.2                      & 62.5                      & 17.7                      & 23.2                     \\
Non Numerical        & 77.2                      & 89.4                      & 23.6                      & 30.9                     \\
Implicit Numerical   & 54.7                      & 50.9                      & 16.7                      & 21.9                     \\
Change with Action   & 66.3                      & 78.2                      & 20.3                      & 26.6                     \\
Change with Time     & 58.3                      & 66.6                      & 17.8                      & 23.4                     \\ \hline
\end{tabular}
}
\caption{
   Category wise \textit{Exact Match Accuracy} of GPT-3 on BCQ and MCQ in one-shot setting. 
    }
    
\label{tab:one_shot_category_results}
\end{table}



\begin{figure}[t!]
    \centering
    \includegraphics[width=7cm]{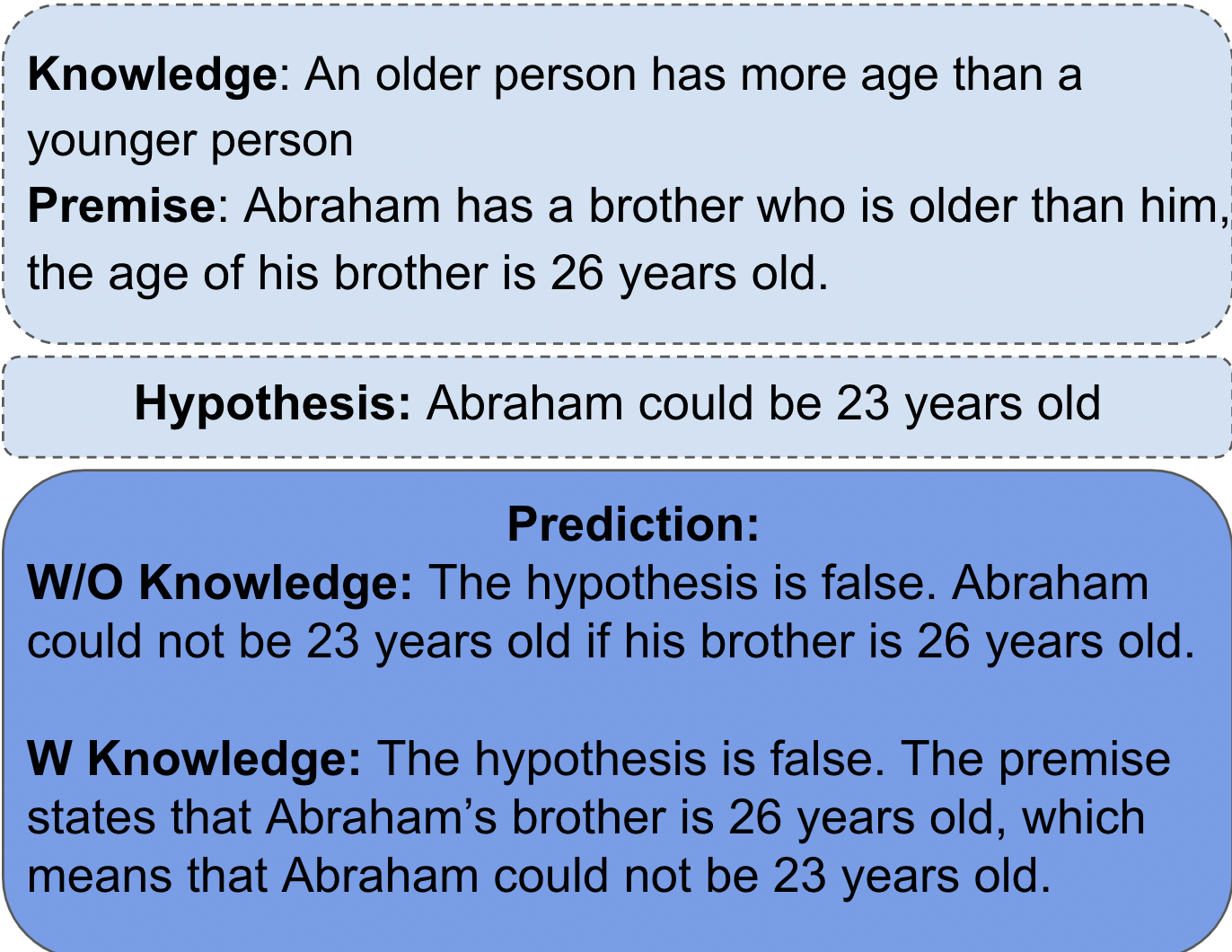}
    \caption{Answers with explanations given by GPT-3 on \name{} dataset.}
    \label{fig:combined example}
\end{figure}

\begin{table}[t!]
\centering
\begin{tabular}{lrr|rr}
\hline
          & \multicolumn{2}{c|}{\textbf{BCQ}}                     & \multicolumn{2}{c}{\textbf{MCQ}}                    \\ \hline
          & \multicolumn{1}{l}{w/o K} & \multicolumn{1}{l|}{w/ K} & \multicolumn{1}{l}{w/o K} & \multicolumn{1}{l}{ w/ K} \\ \hline
Zero Shot & 61.3                      & 70.2                     & 20.1                      &  25.1                    \\
One Shot  & 59.7                      & 67.2                     & 19.5                     & 22.8                    \\
Few Shot  & 65.4                      & 69.1                     & 23.4                     & 35.7                    \\ \hline
\end{tabular}

\caption{
  \textit{Exact Match Accuracy} of GPT-3 on BCQ and MCQ tasks with chain of thought setting.
    }

    
\label{tab:gpt3_w_exp}
\end{table}

\paragraph{Investigating Chain of Thoughts Prompting:}
Recent work has demonstrated the success of learning from instructions \cite{wei2021finetuned, wang2022benchmarking,mishra-etal-2022-cross,mishra-etal-2022-reframing,Lu2022LearnTE,Parmar2022InBoXBARTGI,mishra2022help,luo2022biotabqa} and chain of thought \cite{wei2022chain} and scratchpad prompting \cite{nye2021show}. 
To test this on \name{}, we add explanations to one-shot and few-shot examples provided in the context. 
Table \ref{tab:gpt3_w_exp} shows marginal improvement.
More details are in Appendix \ref{app_c}. 


\paragraph{A Case Study on Prompting GPT-3 to Provide Explanation:} 
We further investigate the reason behind GPT-3's poor performance on \name{} by prompting it to provide the reason behind its answer. Specifically, we add \textit{"Explain the reason behind your answer"} in the prompt. 
Figure \ref{fig:combined example} illustrates a response from GPT-3. 
The answer demonstrates that it did not understand the numerical value of Abraham's age. 
We also provide additional knowledge to assist the model, as shown in Figure \ref{fig:combined example}. 
Even with knowledge, the model could not understand the feasible age. 

\section{Conclusion}
In this work, we proposed \name{}, a question-answering dataset that evaluates the understanding of feasibility. 
We conducted extensive experiments with several state-of-the-art models in zero-shot, one-shot, and few-shot settings and show that these models struggle to answer the feasibility questions correctly. 
We also experimented by providing additional knowledge (relevant to the question) and showed that it leads to a small gain in performance, but the overall performance still remains low.
We further analyzed the performance of models that reveals several interesting findings.
Finally, we release our dataset and hope that our work will encourage further research in feasibility reasoning, an important yet underexplored aspect of commonsense reasoning.

\section*{Limitations}
Like any other commonsense reasoning ability, the concept of feasibility, in general, can be applied in numerous real-world situations.
In our dataset, we try to cover a diverse set of such situations that test the understanding of feasibility, but it is in not an exhaustive list. 
In the future, we will expand the category space by either converting existing numerical datasets into feasibility questions or manually creating new category spaces. 
Along with the dataset, we release the list of contexts and situations on which the question is based. 
In the future, this would help expand the dataset to cover other domains and situations.
The human evaluation of the dataset could also be an interesting study, but it can be an expensive. 
The selection of humans in terms of their educational background and age is also required for unbiased evaluation. 
A completely random selection of humans is also required for a comprehensive study. 
Finally, our dataset includes questions in only one language, i.e., English.



\section*{Ethical Considerations}

The names used in this dataset are selected from the most common English names. 
In question creation, we ensure that all our contexts and questions describe realistic situations, e.g., we keep a range of numerical entities and units appropriate for their context.
No personal information from data creators has been collected during the creation of the dataset.  

\bibliographystyle{acl_natbib}
\bibliography{anthology,custom}

\appendix
\section*{Appendix}

\section{Dataset characteristics }
\label{appendixa_dataset}

In this section, we describe \name{} in more detail. Table \ref{tab:dataset_example} shows illustrative examples of each category discussed in section 4. Each row of the 4th Column of the table shows three questions that were prepared in response to a context. Table \ref{tab:dataset_distrib} gives the distribution of each category of the dataset.  We will explain the motivation behind each category. Please note that the explanations are with respect to examples presented in Table 
 \ref{tab:dataset_example}.

\textbf{Attribute Comparison} shows the comparative properties between two similar objects. The context from attribute comparison is designed to show that quantities can be measured using words like higher and lower, and the model has to understand the relation between them to answer different questions. In this example, it is not possible that the smaller bottle can have a volume of 33 units since the larger one is 32 units. 

\textbf{Change with Time} gives the events that have time as the changing factor. The context is designed to test the model's ability to deduce time-based changes and how certain actions/ events/ quantities can or cannot be done before/ after a certain time. In this case, it is impossible that Edward's age could be 19 on his last birthday as his current age is 16.

\textbf{Change with Action} describes the actions which alter certain quantities/events and test the model's ability to understand that. In this case, it is demonstrated that selling/giving away a certain quantity reduces it. In the example, it is demonstrated that selling all 273 items at least 1 dollar will leave Joshua with at least 273 dollars. Hence the question that he could have 260 dollars is false. 

\textbf{Implicit Numerical Knowledge} tests the model's ability to understand numerical entities as facts and how to manipulate them in different situations. In this case, using the knowledge (or without using it) that four quarters make 1 dollar, the model needs to understand how many quarters will be used in 12 dollars, which is 48 quarters. Hence the question tells us that Christopher can have 52 quarters. 

\textbf{Non Numerical} category tests the model's understanding of very broad domains. They do not have to be numerical based in all the cases. 

The dataset contains diverse topics ranging from physics, mathematics, biology, and numerical reasoning. A total of 422 subcategories are present in the dataset.
Table \ref{tab:dataset_distrib} shows the distribution of BCQ and MCQ questions across different categories in the dataset.

\section{Other performance results }
\label{appendixb_performance_results}

\subsection{Performance Metrics}
For the MCQ setting of the dataset, we study true positive, false positive, and false negative rates as the evaluation metrics. Each metric definition is listed below: 

    
    \textbf{False Negative rate} is defined as the number of incorrect predictions the model gave as correct. For example, if the model gave output as A, B, C, and the predicted result is A, C, then B is missed. The number of false negatives would be 1 (B).
    
    \textbf{False Positive rate} is defined as the number of correct predictions the model gave as incorrect. For example, if the given output is A, B and the predicted result is A, B, C, then the number of false negatives would be 1 (C).

\begin{table}[t!]
\centering
\begin{tabular}{lrr}
\hline
\textbf{} & \multicolumn{1}{l}{w/o  Knowledge} & \multicolumn{1}{l}{w/  Knowledge} \\ \hline
Zero      & 0.17                                   & 0.13                                \\
One       & 0.36                                   & 0.32                                \\
Few       & 0.33                                   & 0.24                                \\ \hline
\end{tabular}

\caption{
   False Positive rate of GPT-3 on MCQ section 
    }
    
\label{tab:false_positive_rate}
\end{table}
\begin{table}[t!]
\centering
\begin{tabular}{lrr}
\hline
\textbf{} & \multicolumn{1}{l}{w/o  Knowledge} & \multicolumn{1}{l}{w/  Knowledge} \\ \hline
Zero      & 0.42                                   & 0.42                                \\
One       & 0.21                                   & 0.24                                \\
Few       & 0.18                                   & 0.20                                 \\ \hline
\end{tabular}

\caption{
    False negative rate of GPT-3 on MCQ section 
    }
    
\label{tab:false_negative_rate}
\end{table}


\subsection{Results}

False positive results shown in Table \ref{tab:false_positive_rate} follow trends similar to accuracy where the performance of one-shot experiments is worse than zero-shot and few-shot. But with the addition of knowledge, the false positive rate decreases. 

As shown in Table \ref{tab:false_negative_rate}, the False negative rate decreases from zero-shot to few-shot experiments, but contrary to other experiments, it increases with the addition of knowledge in almost all the cases. 

\begin{table}[t!]
\resizebox{\linewidth}{!}{
\begin{tabular}{lrr}
\hline
\multicolumn{3}{c}{\textbf{Zero shot BCQ}}                                                     \\ \hline
                     & \multicolumn{1}{l}{w/o Knowledge} & \multicolumn{1}{l}{w/ Knowledge} \\ \hline
Attribute Comparison & 51.2                              & 55.8                              \\
Non Numerical        & 72.7                             & 85.7                              \\
Implicit Numerical   & 52.9                             & 52.0                              \\
Change with Action   & 60.7                             & 65.3                              \\
Change with Time     & 55.5                             & 55.5                              \\ \hline
\multicolumn{3}{c}{\textbf{Zero shot MCQ}}                                                    \\ \hline
                     & \multicolumn{1}{l}{w/o Knowledge} & \multicolumn{1}{l}{w/ Knowledge} \\ \hline
Attribute Comparison & 17.9                             & 20.4                              \\
Non Numerical        & 25.5                             & 29.0                              \\
Implicit Numerical   & 18.5                             & 21.1                              \\
Change with Action   & 21.3                             & 24.2                              \\
Change with Time     & 19.4                             & 22.2                              \\ \hline
\end{tabular}
}

\caption{
    Category wise Accuracy of GPT-3 on BCQ and MCQ task in zero-shot setting.
    }
    
\label{tab:zero_shot_category_results}
\end{table}
\begin{table}[t!]
\centering
\resizebox{\linewidth}{!}{
\begin{tabular}{lrr}
\hline
\multicolumn{3}{c}{\textbf{Few shot BCQ}}                                                      \\ \hline
                     & \multicolumn{1}{l}{w/o Knowledge} & \multicolumn{1}{l}{w/ Knowledge} \\ \hline
Attribute Comparison & 64.5                             & 69.5                              \\
Non Numerical        & 85.9                             & 99.4                              \\
Implicit Numerical   & 60.8                             & 56.6                              \\
Change with Action   & 73.7                             & 86.8                              \\
Change with Time     & 64.8                             & 74.1                              \\ \hline
\multicolumn{3}{c}{\textbf{Few shot MCQ}}                                                     \\ \hline
                     & \multicolumn{1}{l}{w/o Knowledge} & \multicolumn{1}{l}{w/ Knowledge} \\ \hline
Attribute Comparison & 25.2                             & 35.8                              \\
Non Numerical        & 33.5                             & 47.6                              \\
Implicit Numerical   & 23.7                             & 33.7                              \\
Change with Action   & 28.8                             & 40.9                              \\
Change with Time     & 25.3                             & 35.9                              \\ \hline
\end{tabular}
}

\caption{
    Category wise Accuracy of GPT-3 on BC and MCQ task in few-shot setting.
    }
    
\label{tab:few_shot_category_results}
\end{table}

Table \ref{tab:zero_shot_category_results} shows the category-wise results in zero-shot settings for BCQ and MCQ tasks. For the BCQ task, accuracy was lowest in the Attribute comparison category and highest in Non-Numerical Category. Performance of the Non Numerical category improved significantly in with knowledge setting.

In the MCQ portion of the dataset, the performance gap between Non-Numerical and other categories reduces significantly. It is still the best-performing category for the model, but the Change with Action Category also produced similar results. There was no significant improvement in both Non-Numerical and change with action as observed in the Non-Numerical with the addition of knowledge.

Table \ref{tab:few_shot_category_results} shows the category-wise results for BCQ and MCQ tasks in few shot setting. For the BCQ task, accuracy was lowest in the Attribute comparison category and highest in Non-Numerical Category. Performance of the Non Numerical category improved significantly in the knowledge setting with accuracy reaching above 90\% for the first time in any of the categories.

In the MCQ portion of the dataset, the performance gap between Non-Numerical and other categories reduces significantly. It is still the best-performing category for the model. There was a significant improvement in  Non Numerical and change with action and change with time categories with the addition of knowledge.

\paragraph{Exact 50\% accuracy of GPT-2:} 
\label{appendixb1_performance_results} The input format for all models was as follows: Zero-Shot, Question (Different questions), and Options (True or False). Example(s) preceded the question in the one-shot and few-shot settings. Based on this format, GPT-2 gave the probability of “False” higher in all cases. Since the dataset is label balanced, all the True hypothesis questions were incorrectly predicted, hence giving a 50\% accuracy.


\begin{table*}[t!]
    \centering
    \small
    \resizebox{\linewidth}{!}
    {
    \begin{tabular}{lll}
\hline
\textbf{\begin{tabular}[c]{@{}l@{}}Example or\\ Evaluation sample\end{tabular}}                                &                                                                  & \textbf{Text / Context}                                                                                                                                                                                                                                       \\ \hline
                                                                                                               & \textbf{Prompt}                                                  & \begin{tabular}[c]{@{}l@{}}Select the best option for the questions based on the Context and explain \\ the reasoning. An Example to show how it works:\end{tabular}                                                                                          \\
\multirow{4}{*}{\begin{tabular}[c]{@{}l@{}}Example fed in \\ GPT-3 as 1 shot \\ learning example\end{tabular}} & \multicolumn{1}{c}{\textbf{Context}}                             & \begin{tabular}[c]{@{}l@{}}Eric boiled a glass of water to drink the hot water. The level of the water in the \\ glass is 63 units. After boiling was done, she took the water into the glass.\end{tabular}                                                   \\
                                                                                                               & \textbf{Question}                                                & What could be the level of the water after boiling?                                                                                                                                                                                                           \\
                                                                                                               & \multicolumn{1}{c}{\textbf{Options}}                             & A: 65, \textbf{B: 59}, C: 64, \textbf{D: 61}, E: None                                                                                                                                                                                                                                    \\
                                                                                                               & \textbf{Answer:}                                                 & \begin{tabular}[c]{@{}l@{}}Since boiling leads to evaporation, there is a possibility that the volume of water \\ would have decreased. Due to this, the current water level should be less than \\ 63 units. Hence the answer is B:59 and D: 61\end{tabular} \\
                                                                                                               &                                                                  &                                                                                                                                                                                                                                                               \\
\multirow{4}{*}{\begin{tabular}[c]{@{}l@{}}Example given to\\ GPT-3 for \\ evaluation\end{tabular}}            & \multicolumn{1}{c}{\textbf{Context}}                             & A negative number was added to 70.                                                                                                                                                                                                                            \\
                                                                                                               & \textbf{Question}                                                & What could be the result?                                                                                                                                                                                                                                     \\
                                                                                                               & \multicolumn{1}{c}{\textbf{Options}}                             & A: 70, B: 89, C: 87, \textbf{D: 68}, E: None                                                                                                                                                                                                                             \\
                                                                                                               & \textbf{\begin{tabular}[c]{@{}l@{}}GPT-3\\ Answer\end{tabular}}  & \begin{tabular}[c]{@{}l@{}}The result could be A: 70, since \textbf{adding a negative number would not change the} \\ \textbf{value of 70}. The result could also be C: 87, since adding a negative number \\ would increase the value of 70\end{tabular}                       \\
                                                                                                               &                                                                  &                                                                                                                                                                                                                                                               \\
\multirow{4}{*}{\begin{tabular}[c]{@{}l@{}}Example given to\\ GPT-3 for \\ evaluation\end{tabular}}            & \multicolumn{1}{c}{\textbf{Context}}                             & \begin{tabular}[c]{@{}l@{}}Last year 758 people joined the company and this vear the number of people \\ who joined was more than double the last time\end{tabular}                                                                                           \\
                                                                                                               & \textbf{Question}                                                & How many people could have joined this year?                                                                                                                                                                                                                  \\
                                                                                                               & \multicolumn{1}{c}{\textbf{Options}}                             & A: 1440, \textbf{B: 1600}, C: 1460, D: 1370, E: None                                                                                                                                                                                                                   \\
                                                                                                               & \textbf{\begin{tabular}[c]{@{}l@{}}GPT-3\\ Answer:\end{tabular}} & \begin{tabular}[c]{@{}l@{}}Since the number of people who joined this year was \textbf{more than double the} \\ \textbf{last time}, the number of people who joined this year could be up to 1460. \\ Therefore, the answer is C: 1460.\end{tabular}                            \\ \hline
\end{tabular}
    }
    \caption{
     Illustrating chain of thought approach on some examples of feasibilityQA dataset in 1 shot setting without providing knowledge. 1st set of rows demonstrate the example fed into GPT-3 for 1 shot learning. 2nd and 3rd set of rows show GPT-3's response to Context, Question and Options asked.
    }
    \label{tab:case_study_chain_of_command}
\end{table*}

\begin{table*}[t!]
    \centering
    \small
    \resizebox{\linewidth}{!}
    {
        \begin{tabular}{lll}
\hline
\textbf{\begin{tabular}[c]{@{}l@{}}Example or\\ Evaluation sample\end{tabular}}                                &                                                                 & \textbf{Text / Context}                                                                                                                                                                                                                                       \\ \hline
                                                                                                               & \textbf{Prompt}                                                 & \begin{tabular}[c]{@{}l@{}}Select the best option for the questions based on the context, knowledge \\ and explain the reasoning. An Example to show how it works:\end{tabular}                                                                               \\
\multirow{5}{*}{\begin{tabular}[c]{@{}l@{}}Example fed in \\ GPT-3 as 1 shot \\ learning example\end{tabular}} & \textbf{Context}                                                & \begin{tabular}[c]{@{}l@{}}Eric boiled a glass of water to drink the hot water. The level of the water in the \\ glass is 63 units. After boiling was done, she took the water into the glass.\end{tabular}                                                   \\
                                                                                                               & \textbf{Knowledge}                                              & Water will convert into water vapour on boiling.                                                                                                                                                                                                              \\
                                                                                                               & \textbf{Question}                                               & What could be the level of the water after boiling?                                                                                                                                                                                                           \\
                                                                                                               & \textbf{Options}                                                & A: 65, B: 59, C: 64, D: 61                                                                                                                                                                                                                                    \\
                                                                                                               & \textbf{Answer}                                                 & \begin{tabular}[c]{@{}l@{}}Since boiling leads to evaporation, there is a possibility that the volume of water \\ would have decreased. Due to this, the current water level should be less than \\ 63 units. Hence the answer is B:59 and D: 61\end{tabular} \\
                                                                                                               &                                                                 &                                                                                                                                                                                                                                                               \\
\multirow{5}{*}{\begin{tabular}[c]{@{}l@{}}Example given to\\ GPT-3 for \\ evaluation\end{tabular}}            & \textbf{Context}                                                & Two boys competed in a race. The loser finished the race in 24 minutes.                                                                                                                                                                                       \\
                                                                                                               & \textbf{Knowledge}                                              & Loser will take more time to finish the race.                                                                                                                                                                                                                 \\
                                                                                                               & \textbf{Question}                                               & How many minutes the other boy could have taken to finish the race?                                                                                                                                                                                           \\
                                                                                                               & \textbf{Options}                                                & A: 32, B: 25, C: 15, D: 22, E: None                                                                                                                                                                                                                           \\
                                                                                                               & \textbf{\begin{tabular}[c]{@{}l@{}}GPT-3\\ Answer\end{tabular}} & \begin{tabular}[c]{@{}l@{}}The other boy \textbf{could have taken less time to finish the race}. \\ Hence, the answer is B: 25.\end{tabular}                                                                                                                           \\ \hline
\end{tabular}
    }
    \caption{
     Illustrating chain of thought approach on some examples of feasibilityQA dataset in 1 shot setting with providing knowledge. 1st set of rows demonstrate the example fed into GPT-3 for 1 shot learning. 2nd row shows GPT-3's response to Context, Question and Options asked.
    }
    \label{tab:case_study_chain_of_command_WK}
\end{table*}

\section{Case study: Chain of Thought Reasoning Approach}
\label{app_c}
Table \ref{tab:case_study_chain_of_command} and \ref{tab:case_study_chain_of_command_WK} show the unsuccessful attempts in the chain of thought reasoning approach. Table \ref{tab:case_study_chain_of_command} shows the setting where the 1st example is fed into the model as an example of how to reason out the answer. The reason and answer were clearly mentioned that told that evaporation leads to a decrease in water level and hence water level should decrease. This led to a decrease in water level; hence, the correct answers were quantities less than 63; \textbf{59 and 61}. 

The 2nd and 3rd sets of rows show the Context, question, and options supplied to GPT-3, and we get responses in GPT-3 Answer row. The logic given for the addition of a number is wrong. Adding a negative number should decrease the value, and hence rest of the answer will be wrong. In the 3rd row GPT-3's response, the logic used to calculate the answer is correct, but it was unable to calculate that 1600 was double 758. Both parts are highlighted in the table.

The situation did not improve much when knowledge was supplied with other rows, as shown in Table \ref{tab:case_study_chain_of_command_WK}. The model was able to interpret the logic correctly but could not associate that logic with numerical quantities.

\end{document}